\documentclass{article}

\usepackage[final]{neurips_2023}
\usepackage{amsmath}
\usepackage{graphicx} 

\usepackage[utf8]{inputenc} 
\usepackage[T1]{fontenc}    
\usepackage{hyperref}       
\usepackage{url}            
\usepackage{booktabs}       
\usepackage{amsfonts}       
\usepackage{nicefrac}       
\usepackage{microtype}      
\usepackage{xcolor}         

\title{Athena: Efficient Block-Wise Post-Training Quantization for Large Language Models Using Second-Order Matrix Derivative Information}

\author{%
  Yanshu Wang\\
  Peking University\\
  Beijing, China\\
  \texttt{yanshuwang@pku.edu.cn} \\
  \And
  Wenyang He \\
  Peking University \\
  Beijing, China\\
  \texttt{hwy@stu.pku.edu.cn} \\
  \And
  Tong Yang \\
  Peking University \\
  Beijing, China\\
  \texttt{yangtong@pku.edu.cn} \\
}

\begin{document}

\maketitle

\begin{abstract}
Large Language Models (LLMs) have significantly advanced natural language processing tasks such as machine translation, text generation, and sentiment analysis. However, their large size, often consisting of billions of parameters, poses challenges for storage, computation, and deployment, particularly in resource-constrained environments like mobile devices and edge computing platforms.

Effective compression and quantization techniques are crucial for addressing these issues, reducing memory footprint and computational requirements without significantly compromising performance. Traditional methods that uniformly map parameters to compressed spaces fail to account for the uneven distribution of parameters, leading to substantial accuracy loss.

In this work, we propose Athena, a novel algorithm for efficient block-wise post-training quantization of LLMs. Athena leverages Second-Order Matrix Derivative Information to guide the quantization process using the curvature information of the loss landscape. By grouping parameters by columns or rows and iteratively optimizing the quantization process, Athena updates the model parameters and Hessian matrix to achieve significant compression while maintaining high accuracy. This makes Athena a practical solution for deploying LLMs in various settings.

\end{abstract}

\section{Introduction}


Large Language Models (LLMs) have revolutionized the field of natural language processing, enabling significant advancements in tasks such as machine translation, text generation, and sentiment analysis. Despite their impressive capabilities, the sheer size of these models, which often consist of billions of parameters, presents substantial challenges in terms of storage, computation, and deployment. These challenges are particularly pronounced in resource-constrained environments, such as mobile devices and edge computing platforms.


To address the issues of storage and computational efficiency, it is crucial to develop techniques that can effectively compress and quantize these large models without significantly compromising their performance. Quantization reduces the precision of the model parameters, thereby decreasing the memory footprint and computational requirements. Effective quantization techniques can make it feasible to deploy LLMs in a wider range of applications and devices, broadening their accessibility and utility.

In practice, when compressing parameters, the distribution of parameters is often uneven. Previous algorithms uniformly and linearly map parameters to another compressed space. Although this method is simple, it may not be effective and can result in significant accuracy loss because it does not take into account the original data distribution. It maps both densely populated and sparse regions to the quantized space in the same way.

\begin{figure}
\centering
\includegraphics[width=9.2cm]{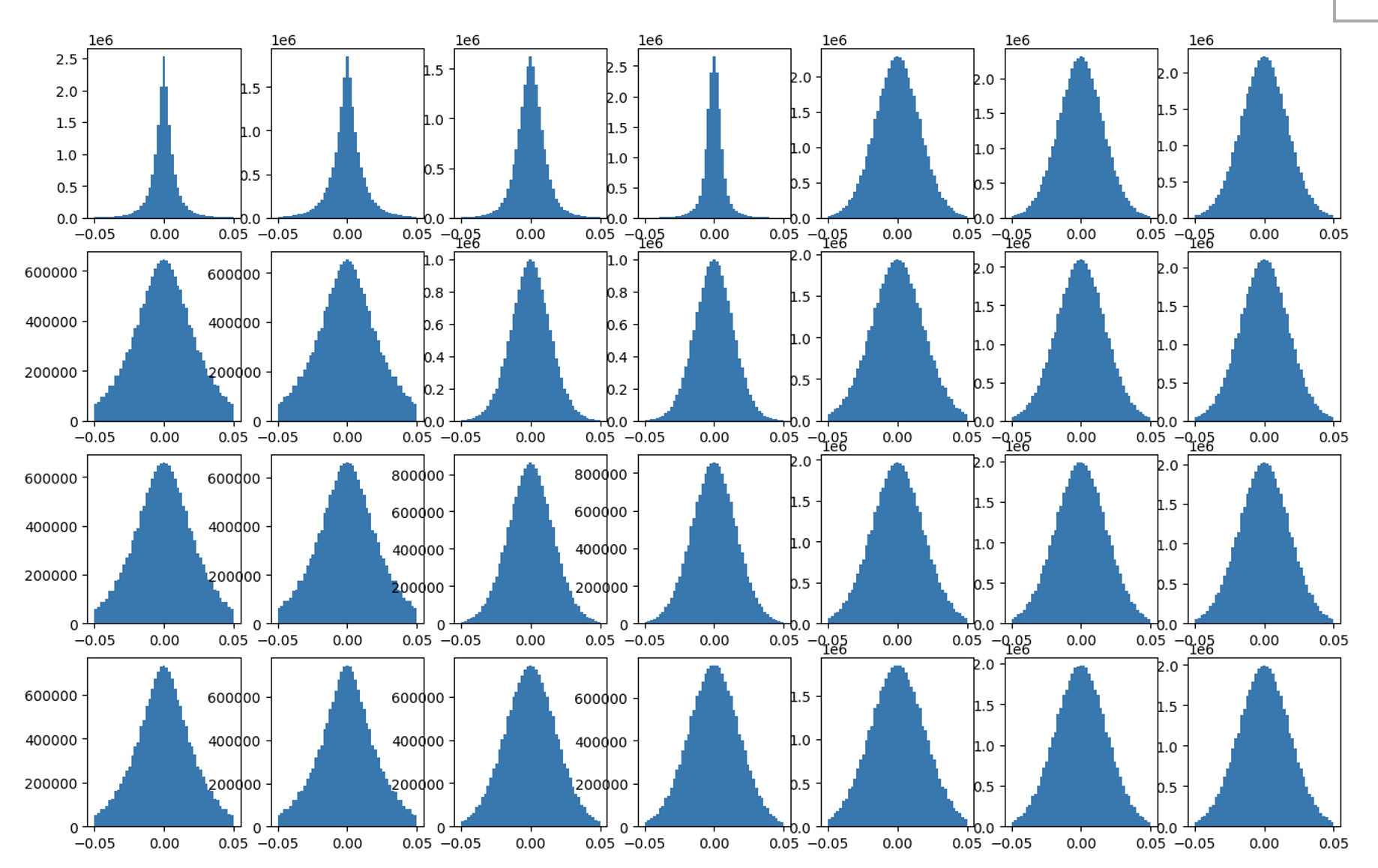}
\caption{The distribution of parameter. }
\label{fig:dis}
\end{figure}

A lines of works have been proposed to quant the model weights~\cite{dettmers2208llm,polino2018model,chmiel2020robust,fan2020training,zafrir2019q8bert,wu2022xtc,lecun1989optimal}.The distribution of Transformer neural network weights is shown in Figure ~\ref{fig:dis}. In Figure ~\ref{fig:dis}, different rows represent the 0th layer, 8th layer, 16th layer, and 24th layer. Different columns represent the attention layer's q, k, v, and o layers, as well as the fully connected layer's gate, up, and down layers. The distributions of different layers are not uniformly analyzed, and the distributions vary across different layers. Therefore, a more reasonable approach is to quantize the model based on the importance of the parameters, stratifying and computing accordingly.


In this work, we propose a novel algorithm named Athena to achieve efficient block-wise post-training quantization for large language models. The core idea of Athena is to utilize Second-Order Matrix Derivative Information, which leverages the curvature information of the loss landscape to guide the quantization process. This approach ensures that the model compression is done in a way that minimally impacts the model's performance.

Athena operates by grouping model parameters by columns or rows to enhance the feasibility of the algorithm. For each group, an iterative quantization process is performed, updating the parameters based on their contribution to the overall model loss. The algorithm iteratively optimizes the quantization process and updates the model parameters and Hessian matrix accordingly. By doing so, Athena achieves significant model compression while maintaining high accuracy, making it a practical solution for deploying LLMs in various settings.


\section{Athena}
\subsection{Second-Order Matrix Derivative Information}

Due to the large size of Large Language Models and the numerous parameters, it is impractical to obtain second-order derivative information for all parameters at once. Moreover, the inference process of the model proceeds layer by layer. We need to compress and optimize based on the performance of each layer. Therefore, the goal of quantizing the parameters is to compress the model as much as possible without affecting the layer-wise loss ($\mathcal{L}$).

Let $\delta \mathbf{w}$ denote a small perturbation and $\mathcal{L}(\mathbf{x}, \mathbf{y}, \mathbf{w})$ denote the task loss that we want to minimize. Then

\begin{align}
    \mathbb{E} &\left[ \mathcal{L} (\mathbf{x}, \mathbf{y}, \mathbf{w} + \delta \mathbf{w}) - \mathcal{L} (\mathbf{x}, \mathbf{y}, \mathbf{w}) \right] \\
    & \approx^{(a)} \mathbb{E} \left[ \delta \mathbf{w}^T \cdot \nabla_{\mathbf{w}} \mathcal{L} (\mathbf{x}, \mathbf{y}, \mathbf{w}) \right. \nonumber \\
    & \left. + \frac{1}{2} \delta \mathbf{w}^T \cdot \nabla_{\mathbf{w}}^2 \mathcal{L} (\mathbf{x}, \mathbf{y}, \mathbf{w}) \cdot \delta \mathbf{w} \right] \\
    & = \delta \mathbf{w}^T \cdot \mathbf{g}^{(\mathbf{w})} + \frac{1}{2} \delta \mathbf{w}^T \cdot \mathbf{H}^{(\mathbf{w})} \cdot \delta \mathbf{w},
\end{align}

where (a) uses the second order Taylor series expansion. $\mathbf{g}^{(\mathbf{w})}$ and $\mathbf{H}^{(\mathbf{w})}$ denote the expected gradient and Hessian of the task loss $\mathcal{L}$ with respect to $\mathbf{w}$, i.e.,

\begin{align}
    \mathbf{g}^{(\mathbf{w})} &= \mathbb{E} \left[ \nabla_{\mathbf{w}} \mathcal{L} (\mathbf{x}, \mathbf{y}, \mathbf{w}) \right] \\
    \mathbf{H}^{(\mathbf{w})} &= \mathbb{E} \left[ \nabla_{\mathbf{w}}^2 \mathcal{L} (\mathbf{x}, \mathbf{y}, \mathbf{w}) \right].
\end{align}

Optimization problem: 
\begin{equation}
\min_{\delta \mathbf{w}} \frac{1}{2} \delta \mathbf{w}^T \mathbf{H} \delta \mathbf{w} \quad \text{s.t.} \quad \mathbb{E}_Q \delta \mathbf{w} + \mathbb{E}_Q \mathbf{w} = \mathbb{E}_Q \cdot \text{quant}(\mathbf{w})
\end{equation}

Construct the Lagrangian function and solve:
\begin{equation}
\mathcal{L} = \frac{1}{2} \delta \mathbf{w}^T \mathbf{H} \delta \mathbf{w} + \lambda \left( \mathbb{E}_Q \delta \mathbf{w} + \mathbb{E}_Q \mathbf{w} - \mathbb{E}_Q \cdot \text{quant}(\mathbf{w}) \right)
\end{equation}

\begin{equation}
\frac{\partial \mathcal{L}}{\partial \delta \mathbf{w}} = 0 , \quad \frac{\partial \mathcal{L}}{\partial \lambda} = 0
\end{equation}

Calculation method for the impact of each weight (Q is a set of weights, $\mathbb{E}_Q$ is $|Q| \times d$):
\begin{equation}
\label{equ:opt}
\mathcal{L} = \frac{1}{2} \left( \mathbb{E}_Q \mathbf{w} - \mathbb{E}_Q \cdot \text{quant}(\mathbf{w}) \right)^T \left( \mathbb{E}_Q \mathbf{H}^{-1} \mathbb{E}_Q^T \right)^{-1} \left( \mathbb{E}_Q \mathbf{w} - \mathbb{E}_Q \cdot \text{quant}(\mathbf{w}) \right)
\end{equation}

Parameter update method:
\begin{equation}
\label{equ:adj}
\delta \mathbf{w} = -\mathbf{H}^{-1} \mathbb{E}_Q^T \left( \mathbb{E}_Q \mathbf{H}^{-1} \mathbb{E}_Q^T \right)^{-1} \left( \mathbb{E}_Q \mathbf{w} - \mathbb{E}_Q \cdot \text{quant}(\mathbf{w}) \right)
\end{equation}

Hessian matrix update method:
\begin{equation}
\mathbf{H}_{-Q}^{-1} = \mathbf{H}^{-1} - \mathbf{H}_{(:,Q)}^{-1} \left( \left[ \mathbf{H}^{-1} \right]_{QQ} \right)^{-1} \mathbf{H}_{(Q,:)}^{-1}
\end{equation}

Thus, we only need to iteratively optimize $\mathcal{L}$ according to equation~\ref{equ:opt} and update $\mathbf{w}$ based on the result of each optimization and equation~\ref{equ:adj}.

\begin{figure}
\centering
\includegraphics[width=9.2cm]{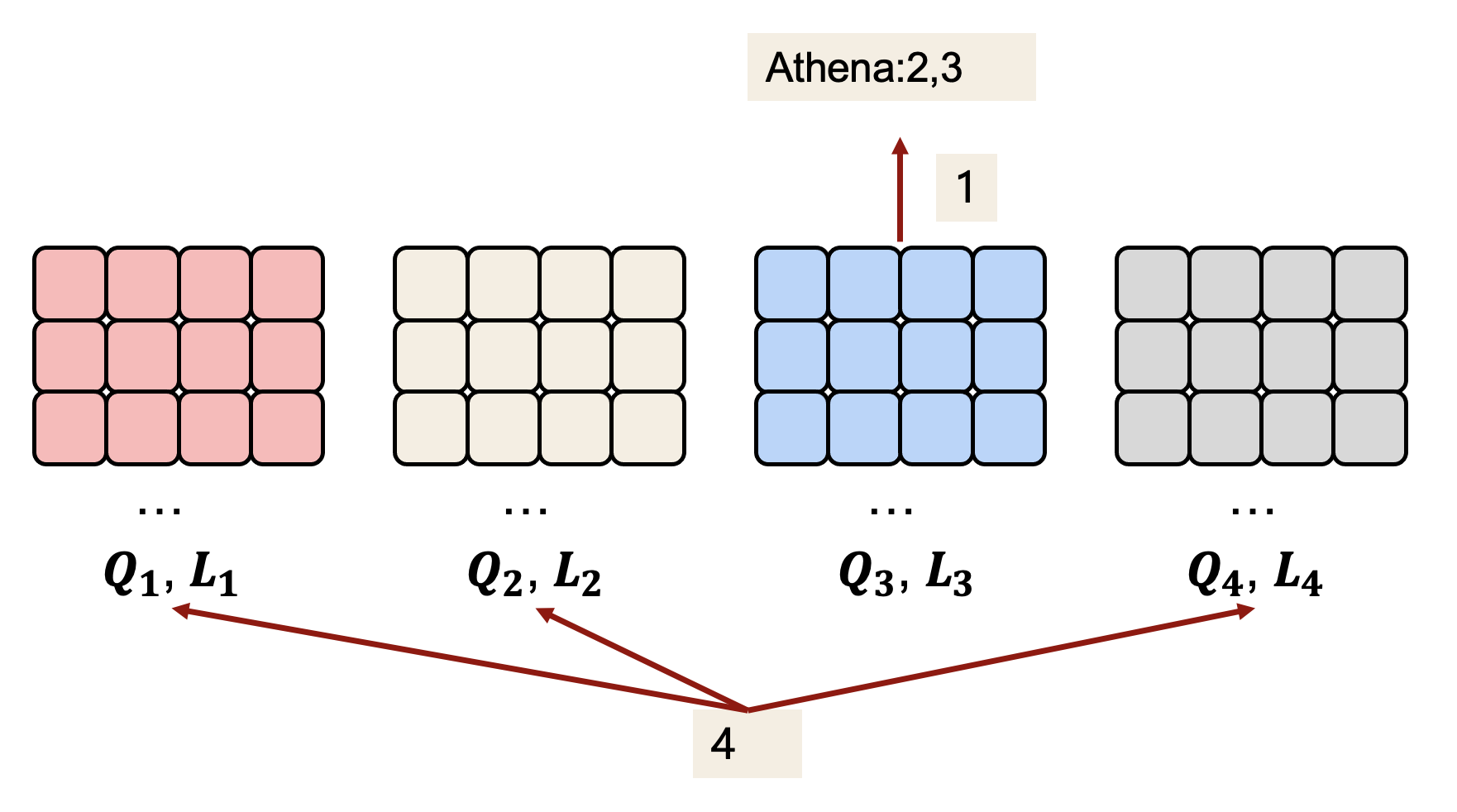}
\caption{The quantization of parameter. }
\label{fig:parquant}
\end{figure}

\subsection{Athena Quantization}
To feasibly apply Second-Order Matrix Derivative Information in model quantization algorithms, we propose Athena. Athena first groups the parameters by columns or rows to enhance the feasibility of the algorithm. Then, for each group, iterative quantization is performed using equation~\ref{equ:opt}. Finally, the parameters $\mathbf{w}$ and the Hessian matrix $\mathbf{H}$ are updated based on the quantization results.

\textbf{Parameter Quantization:}

The Parameter Quantization step is
as shown in figure ~\ref{fig:parquant} step 1$-$3.

We divide the matrix to be compressed into $n$ groups along the column direction, and then perform the following operations:

1. Choose $Q_i$ with the smallest $L$ value based on $L_1, L_2, L_3, L_4$.

2. Use k-means to select the centroids, using $(\mathbb{E}_Q \mathbf{w} - \mathbb{E}_Q \cdot \text{quant}(\mathbf{w}))^T (\mathbb{E}_Q \mathbf{H}^{-1} \mathbb{E}_Q^T)^{-1} (\mathbb{E}_Q \mathbf{w} - \mathbb{E}_Q \cdot \text{quant}(\mathbf{w}))$ as the distance metric instead of Euclidean distance.

3. Encode the vector $Q_3$ based on the centroids.

4. Adjust the weights $Q_1$, $Q_2$, $Q_4$ using $\delta \mathbf{w} = -\mathbf{H}^{-1} \mathbb{E}_Q^T (\mathbb{E}_Q \mathbf{H}^{-1} \mathbb{E}_Q^T)^{-1} (\mathbb{E}_Q \mathbf{w} - \mathbb{E}_Q \cdot \text{quant}(\mathbf{w}))$.

\subsection{Optimisations}

\subsubsection{Optimisation for k-means}
\begin{figure}
\centering
\includegraphics[width=13.2cm]{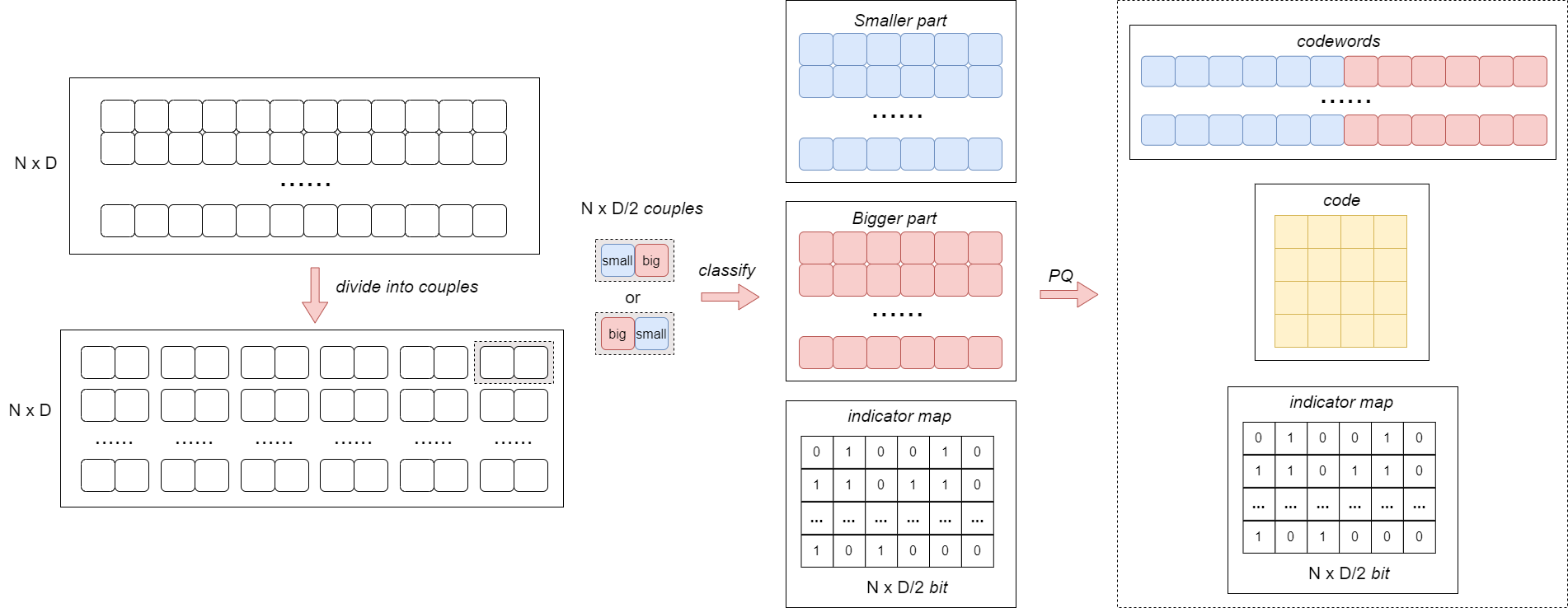}
\caption{Optimisation for k-means. }
\label{fig:kmeamsopt}
\end{figure}

In addition, we employed heuristic methods such as flipping to optimize the efficiency of the k-means algorithm, as shown in Figure ~\ref{fig:kmeamsopt}.

\subsubsection{Codebook Quantization}
Building upon the aforementioned quantization process, this paper proposes codebook quantization, which stores the codebook in a format with fewer bits than fp16. Inspired by, and considering that large language models have almost no precision loss when quantized to 8 bits, we adopt a similar approach to quantize the codebook to 8 bits. Specifically, for each $n \times d$ codebook $C' = C[i, j]$, we take the minimum value $\text{min}_i$ and the maximum value $\text{max}_i$ for the $i$-th dimension of $d$, and map each number to a value between [0, 255]:

\[
C''[a, b] = 255 \cdot \frac{(C'[a, b] - \text{min}_b)}{\text{max}_b}
\]

Each number is then rounded to the nearest integer to get a value in the range [0, 255], which can be stored using 8 bits, thus quantizing the codebook to 8 bits and saving half the codebook storage space, albeit with some precision loss. Notably, in the quantization algorithm, when quantizing $W$ column by column, each column is quantized first, then the codebook is quantized, and the parameters are updated with the quantized codebook. This is in contrast to quantizing the entire model first and then quantizing the codebook, aligning better with the LDLQ algorithm's workflow.

Experiments show that codebook quantization significantly impacts model accuracy, which contrasts sharply with the near-zero precision loss when directly quantizing the model to INT8. This discrepancy may be due to the extensive use of the codebook, which amplifies its impact on the model. The following experimental evaluation section will present a comparison of model accuracy before and after applying codebook quantization.

\subsubsection{Residual Low-Rank Decomposition}
Define the residual $R = \mathbf{W} - \hat{\mathbf{W}} \in \mathbb{R}^{N \times M}$, and consider decomposing the residual into two matrices $R \approx \mathbf{A}\mathbf{B}$, where $\mathbf{A} \in \mathbb{R}^{N \times r}$ and $\mathbf{B} \in \mathbb{R}^{r \times M}$, with $r \ll \min(N, M)$ being a smaller dimension. The goal is to minimize the quantization error by quantizing the weights to $\hat{\mathbf{W}} + \mathbf{A}\mathbf{B}$ after removing the decomposed error. Specifically, we first perform a Cholesky decomposition on the Hessian matrix to obtain $\mathbf{H} = \mathbf{U}\mathbf{U}^T$, which is the product of an upper triangular matrix and its transpose, and rewrite the quantization error as

\[
l = \text{tr}(\mathbf{R}\mathbf{U}(\mathbf{R}\mathbf{U})^T)
\]

Next, we perform a singular value decomposition (SVD) on $\mathbf{R}\mathbf{U}$ and take the largest $r$ components. Suppose the SVD of $\mathbf{R}\mathbf{U}$ yields $\mathbf{R}\mathbf{U} = \mathbf{u}\mathbf{D}\mathbf{v}$, where $\mathbf{u}$ and $\mathbf{v}$ are orthogonal matrices, and $\mathbf{D}$ is a diagonal matrix with singular values sorted in descending order. Let

\[
\mathbf{A} = \mathbf{R}^{-1} \mathbf{u}[:, :d]
\]
\[
\mathbf{B} = \mathbf{D}[:d, :d]\mathbf{v}[:d, :]
\]

In experiments, $r$ is generally set to $\min(M, N)/100$, which is 1\% of the original matrix size. Assuming matrices $\mathbf{A}$ and $\mathbf{B}$ are stored in fp16 format, this adds an additional 0.32 bits per weight (bpw) to the quantization bit number. The above computation process represents the optimal solution for low-rank decomposition with the objective of minimizing quantization error.

\subsubsection{k-v cache compassion}
The importance of k-v cache compression cannot be overstated. In large language models, the size of the k-v cache often exceeds the size of the model itself by several times. This is because the k-v cache stores key and value pairs for each token processed, which grows proportionally with the sequence length and the number of layers in the model. As a result, the k-v cache becomes a significant bottleneck in terms of memory usage, especially during inference on resource-constrained devices such as mobile phones and edge computing platforms. 

Compressing the k-v cache can dramatically reduce the memory footprint, enabling more efficient deployment of large language models in these environments. By reducing the size of the k-v cache without compromising its effectiveness, we can maintain the model's performance while significantly lowering the hardware requirements. This makes advanced language models more accessible and practical for a wider range of applications, promoting their use in real-time and low-latency scenarios.

\subsection{Efficient and scalable implementation}

\begin{figure}
\centering
\includegraphics[width=9.2cm]{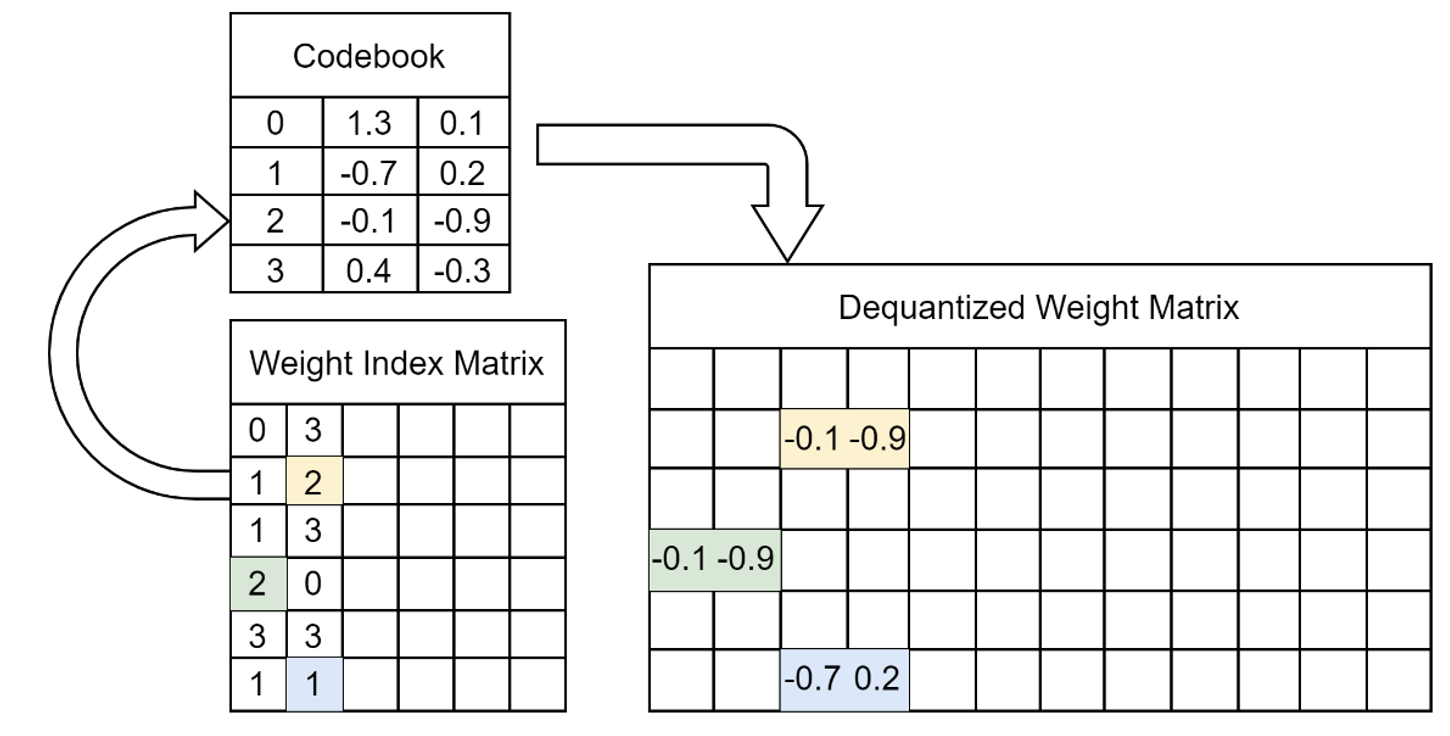}
\caption{The implement of Athena. }
\label{fig:anquant}
\end{figure}
The basic operation of the proposed quantization algorithm is to quantize a weight matrix. Let $\mathbf{W} \in \mathbb{R}^{N \times M}$ be a weight matrix corresponding to the linear function $f(\mathbf{x}) = \mathbf{W}\mathbf{x}$, where $\mathbf{x}$ is any input vector of length $M$. 

The algorithm requires three hyperparameters: $d$ (satisfying $1 \leq d \leq N$), which indicates the dimension of each codebook entry; $n$ (satisfying $1 \leq n \leq k$), which indicates the number of codebook entries, generally a power of 2; and $k$ (satisfying $1 \leq k \leq M$), which indicates the number of points covered by each codebook. The three parameters can be represented in the form $(d, n, k)$, for example, $(2, 64, 1024)$ means that each codebook entry has 2 dimensions, the codebook has 64 elements (with index numbers from 0 to 63), and every 1024 vectors are clustered to generate a codebook.

After quantizing the weight matrix, it is saved as two matrices: an index matrix (Figure ~\ref{fig:anquant} "Weight Index Matrix") and a codebook (corresponding to Figure ~\ref{fig:anquant} "Codebook"). The relationship between these two matrices and the weight matrix is illustrated below.

To intuitively explain the relationship between these two matrices and the original weight matrix, we first introduce the dequantization process. As shown in Figure ~\ref{fig:anquant}, the size of the weight index matrix $I$ is $N \times \lceil M/d \rceil$, where each element is an integer in the range $[0, d-1)$, corresponding to an entry in the codebook and a $1 \times d$ vector in the dequantized matrix. Let the codebook matrix be $C$, whose size is $\lceil N/k \rceil \times \lceil M/d \rceil \times n \times d$, which actually consists of $\lceil N/k \rceil \times \lceil M/d \rceil$ codebooks of size $n \times d$. In the following text, "codebook" refers to a matrix of size $n \times d$.

For the value $I_{(i,j)}$ in the $i$-th row and $j$-th column of the index matrix (starting from 0), the index corresponds to the columns $dj$ to $d(j+1)-1$ in the $i$-th row of the dequantized matrix $\hat{\mathbf{W}}$. Specifically, for $l = 0 \ldots d-1$, we have

\[
\hat{\mathbf{W}}_{(i,dj+l)} = C\left[\left\lceil \frac{i}{k} \right\rceil, j, I_{(i,j)}, l \right]
\]

To optimize the quantization results, this codebook needs to satisfy two properties known as Lloyd's optimality conditions. First, each point must be quantized to the nearest codebook centroid in terms of Euclidean distance:

\[
I_{(i,j)} = \arg \min_{c \in d} \left| C\left[\left\lceil \frac{i}{k} \right\rceil, j, c \right] - \mathbf{W}[i, dj \ldots d(j+1)-1] \right|^2
\]

This ensures that the regions corresponding to each centroid are divided by hyperplanes. Second, the quantization centroids must be the mean of all points in their region, i.e., the centroid.

These two conditions can be ensured iteratively. For an existing set of quantization centroids and index matrix, the first step is to assign the index matrix to the nearest quantization centroid of the corresponding vector in the weight matrix. The second step is to update the quantization centroids to be the mean of all vectors in their region. This iterative process is the k-means clustering algorithm.

\section{Experiments}
\textbf{Setups:} The experiments in this paper were primarily conducted using NVIDIA's RTX 4090 GPU, which has 24GB of memory. Due to memory limitations, we mainly evaluated mainstream models with 7 billion parameters, including Llama-7b\cite{touvron2023llama}, Llama-2-7b\cite{touvron2023llama2}, and Mistral-7b\cite{jiang2023mistral}.

For the calibration dataset, we used C4, which contains approximately 128 segments, each with 2048 tokens. Similar to GPTQ~\cite{frantar2022gptq}, these calibration datasets were used to compute the Hessian matrix. Note that the calibration dataset is independent of the model's actual tasks; the model evaluation is not restricted to the C4 dataset and can be tested on any dataset. 

This paper primarily uses the model's output perplexity to evaluate the performance of the quantized model. The dataset used is WikiText-2\cite{merity2016pointer}, with a context size of 2048, which corresponds to the perplexity\cite{jelinek1977perplexity} when predicting the next token using the first 2047 tokens. Perplexity reflects the model's uncertainty in predicting the next token; the lower the uncertainty, the more accurate the model's predictions, and the lower the perplexity, the smaller the impact of quantization on the model.

In the experiments presented in this paper, there is a good linear relationship between perplexity and quantization precision, making it easy to compare. Therefore, perplexity was chosen as the primary metric for evaluation.

\textbf{Quantization Bit Number}
The quantization bit number refers to the average number of bits used to store each weight. Unlike traditional quantization algorithms, in the proposed algorithm, each weight does not have a fixed bit number. Therefore, a more accurate approach is to calculate the average bit number by dividing the total space occupied after quantization by the number of elements in the weight matrix.

The space occupied after quantization can be divided into two parts. One part is the codebook. Assuming the codebook is stored using 16-bit floating-point numbers, the average space occupied by each weight in the codebook is
\[
b_c = \frac{16n}{k}
\]

The other part is the index. For every $d$ weights, $\lceil \log_2 k \rceil$ bits are used for the index, so the average space occupied by each weight in the index is
\[
b_i = \frac{\lceil \log_2 k \rceil}{d}
\]

Thus, the total quantization bit number is
\[
b = b_c + b_i = \frac{16n}{k} + \frac{\lceil \log_2 k \rceil}{d}
\]

For example, for $(2, 64, 1024)$, we can calculate $b_c = 1$ and $b_i = 3$, so $b = 4$, meaning these hyperparameters correspond to a 4-bit quantization. By adjusting the hyperparameters, this algorithm can be extended to 3-bit, 2-bit, and other quantization schemes, providing an effective quantization solution.


\textbf{Hyperparameters:}Unlike quantization algorithms such as GPTQ that require only one hyperparameter to represent the quantization bit number, the algorithm proposed in this paper requires three hyperparameters: $d$, $n$, and $k$. Here, $d$ satisfies $1 \leq d \leq N$ and represents the dimension of each codebook entry; $n$ satisfies $1 \leq n \leq k$ and represents the number of codebook entries, typically a power of 2; $k$ satisfies $1 \leq k \leq M$ and represents the number of points covered by each codebook. These three parameters can be expressed in the form $(d, n, k)$. For example, $(2, 64, 1024)$ means that each codebook entry has 2 dimensions, the codebook has 64 elements (with index numbers ranging from 0 to 63), and every 1024 vectors are clustered to generate a codebook.



\subsection{Accurate:}
\begin{figure}
\centering
\includegraphics[width=9.2cm]{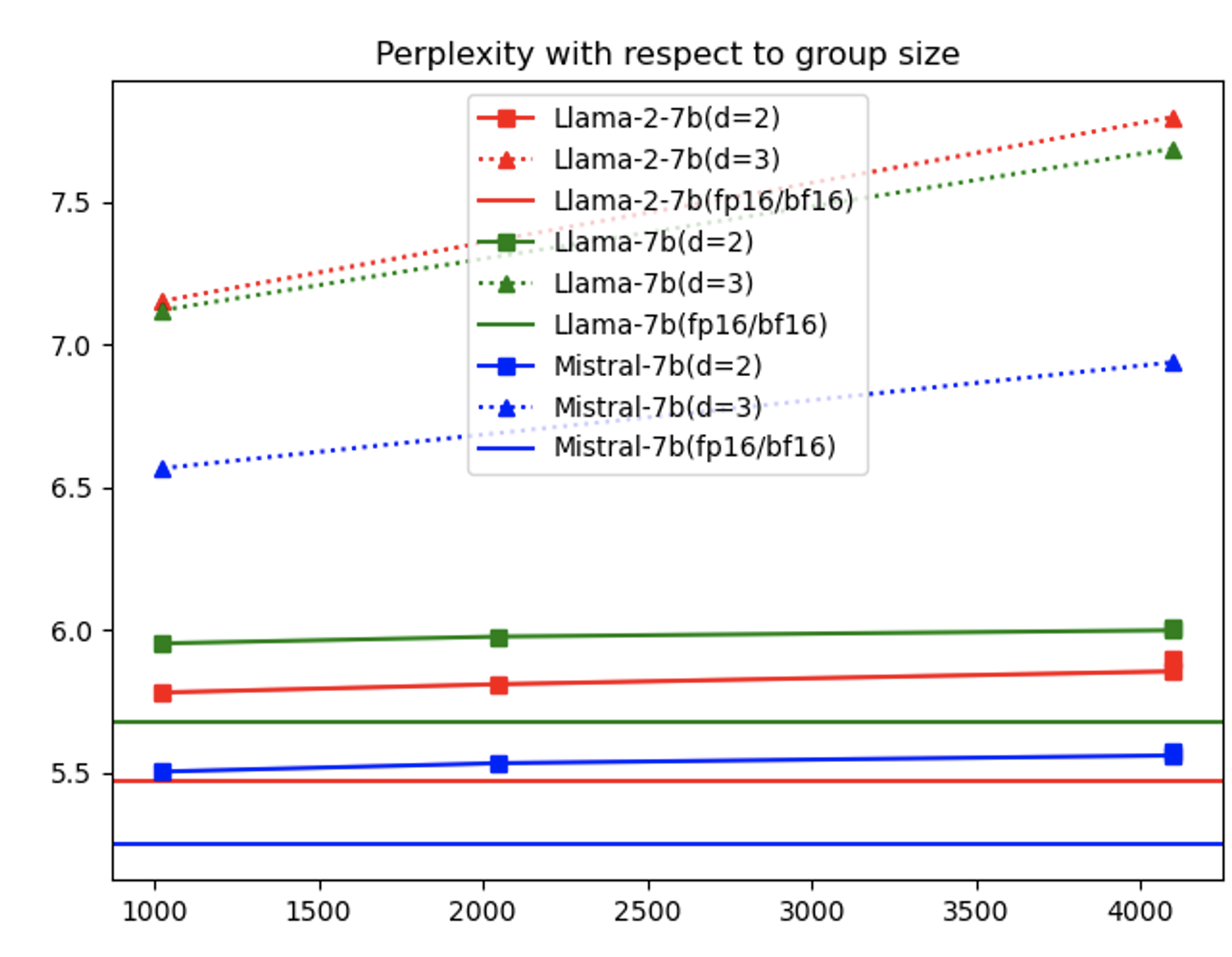}
\caption{ppl vs group sizes. }
\label{fig:pplvsgroupsize}
\end{figure}

As shown in Figure ~\ref{fig:pplvsgroupsize}, the quantization algorithm was run on the Llama-2-7b, Llama-7b, and Mistral-7b models under six different combinations of hyperparameters: $d=2, 3$, $n=64$, and $k=1024, 2048, 4096$. The perplexity was then tested, resulting in the outcomes displayed in the figure. When $d=2$, $b_i=3$, meaning the index of each parameter occupies 3 bits on average, resulting in lower perplexity. Conversely, when $d=3$, $b_i=2$, meaning the index of each parameter occupies 2 bits on average, resulting in higher perplexity.

It can be observed that when $b_i=3$, the actual quantization bit number is close to 3 bits, and the perplexity is low, with $k$ (the group size) having a smaller impact on perplexity. Since a larger $k$ results in a smaller actual quantization bit number with almost no loss in precision, it is suggested that $k$ should be as large as possible. When $b_i=2$, the perplexity is higher, and $k$ has a greater impact on precision.

\begin{figure}
\centering
\includegraphics[width=9.2cm]{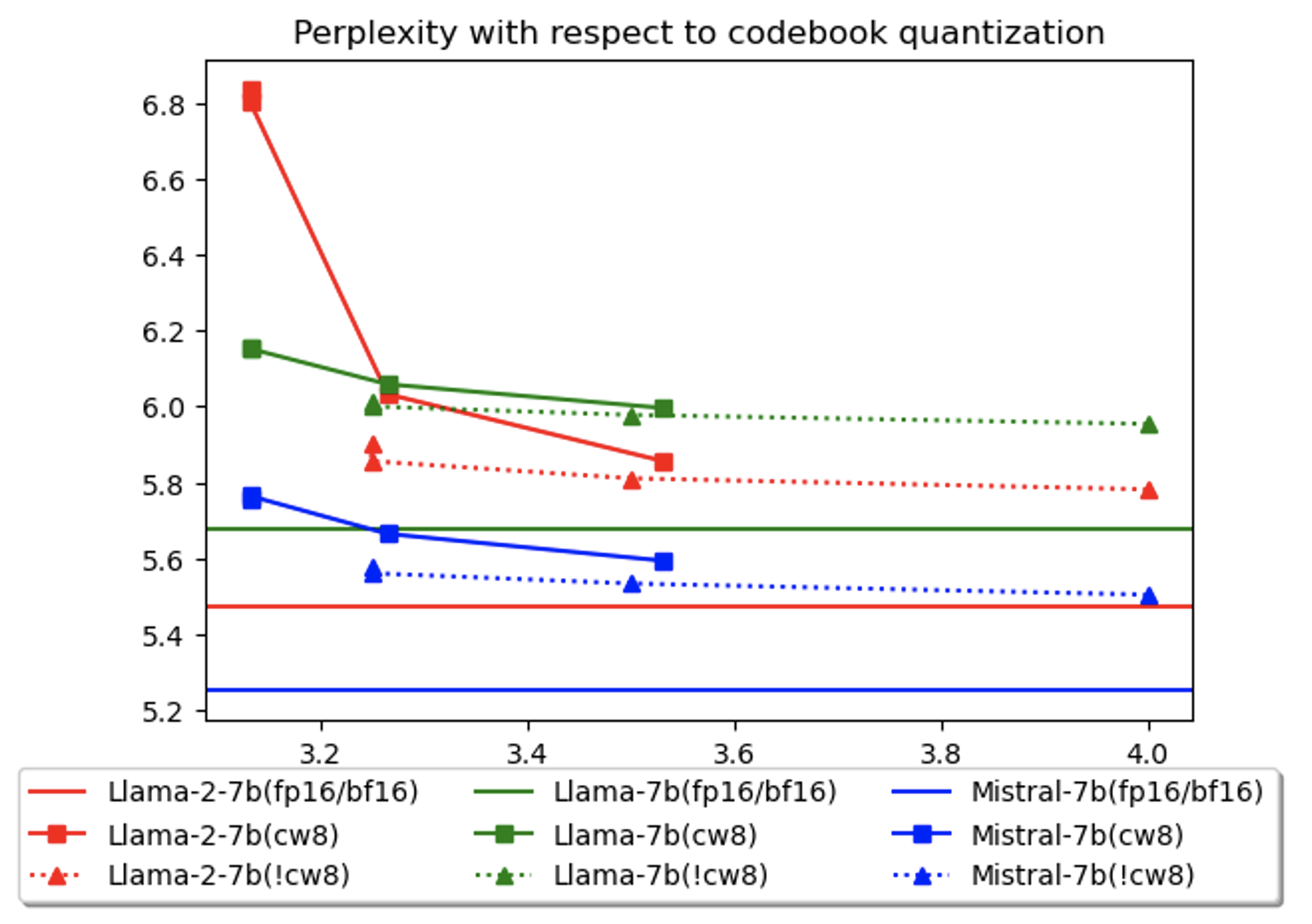}
\caption{ppl using codebook. }
\label{fig:pplvscodebooksize}
\end{figure}

Based on the codebook quantization optimization technique, the effects of the algorithm before and after optimization are compared, as shown in Figure ~\ref{fig:pplvscodebooksize}. The horizontal axis represents the actual quantization bit number. The dashed line indicates the quantization effect without codebook quantization, while the solid line indicates the quantization effect with codebook quantization. 

Because the codebook, originally stored in fp16 format, is converted to int8 format after codebook quantization, the space occupied by the codebook is halved. This results in a lower quantization bit number for the solid line compared to the dashed line under the same hyperparameters, corresponding to higher perplexity.

\begin{figure}
\centering
\includegraphics[width=9.2cm]{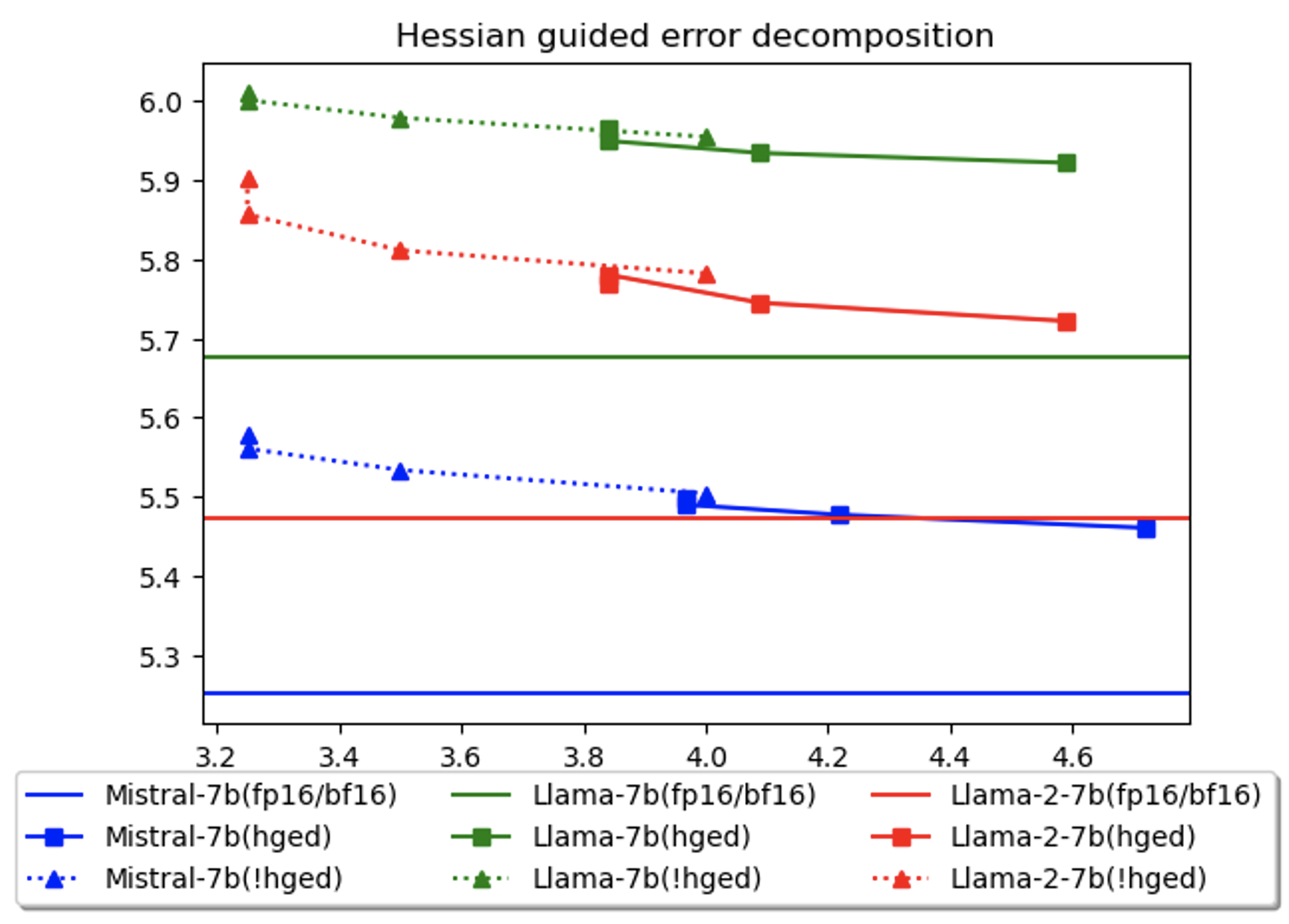}
\caption{ppl using Hessian guide. }
\label{fig:haerror}
\end{figure}

The residual low-rank decomposition can be applied to further reduce the quantization error of the weights by decomposing and storing the error in a low-rank format. As shown in Figure~\ref{fig:haerror}, there is a slight decrease in perplexity, and a corresponding slight increase in the model bit number. The low-rank decomposition occupies approximately 0.32 bits, which can be seen as the solid line shifting to the right compared to the dashed line in the figure.

The residual low-rank decomposition effectively improves the model's accuracy, but there is still a gap compared to the unquantized baseline model. We speculate that this indicates the error distribution is relatively uniform in all directions, making it challenging to capture the majority of the error through low-rank decomposition.

\begin{figure}
\centering
\includegraphics[width=8.2cm]{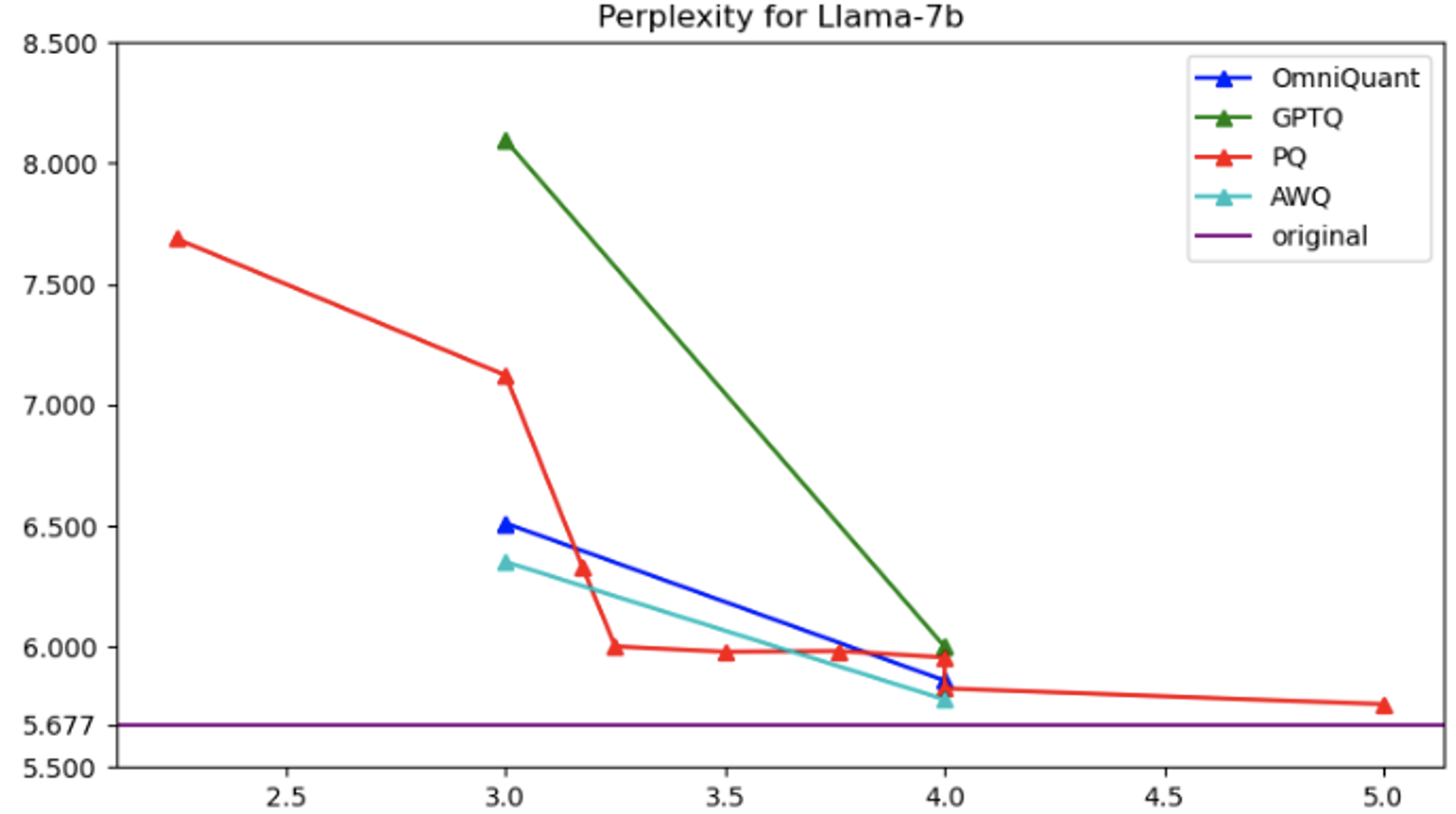}
\caption{Compared to other methods. }
\label{fig:com1}
\end{figure}

\begin{figure}
\centering
\includegraphics[width=8.2cm]{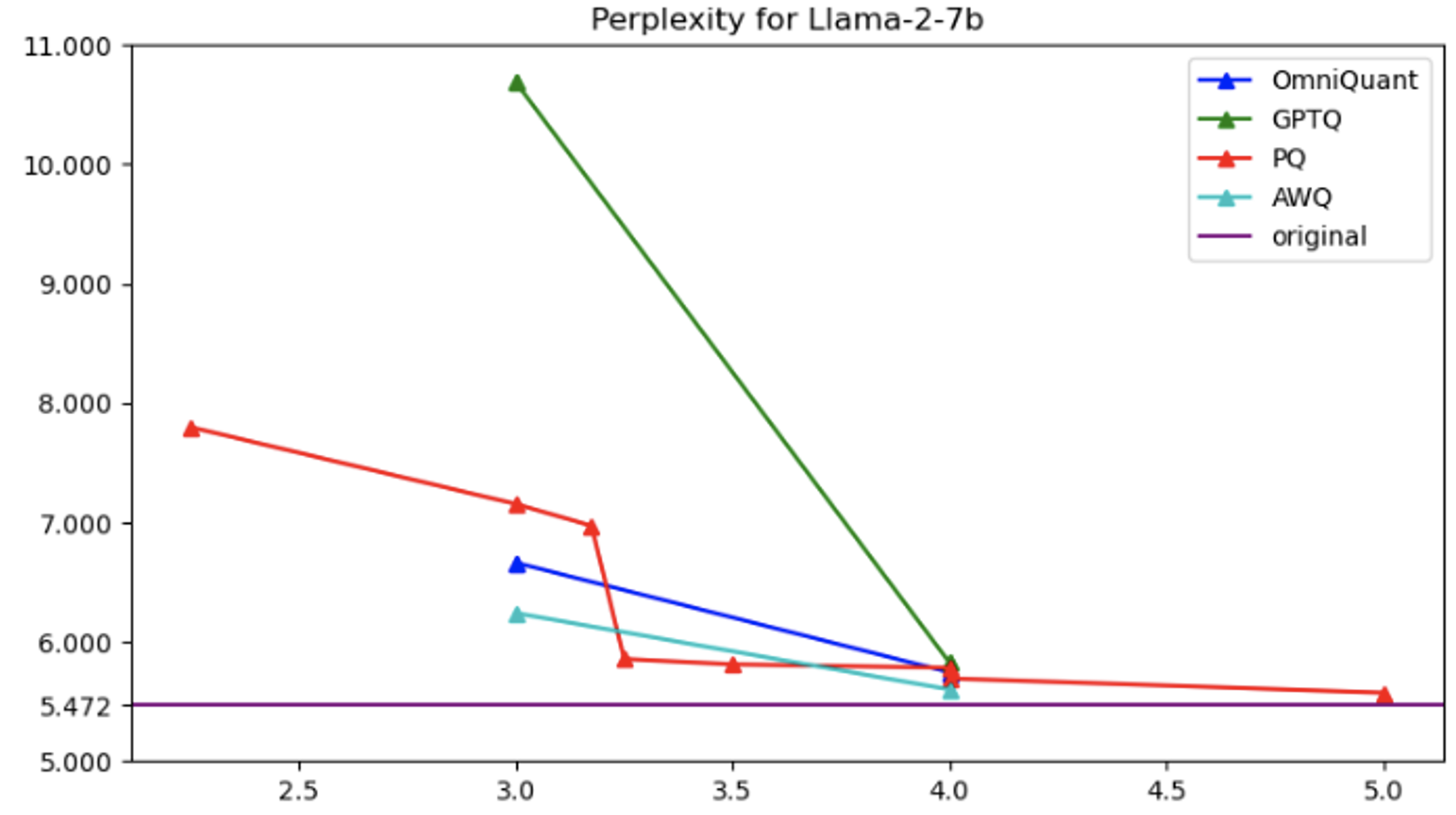}
\caption{Compared to other methods. }
\label{fig:com2}
\end{figure}

To compare with other algorithms, we selected mainstream quantization algorithms such as GPTQ , AWQ\cite{lin2023awq} , and OmniQuant\cite{shao2023omniquant} , and evaluated the perplexity of the Llama-7b and Llama-2-7b open-source models using quantization precision as the horizontal axis. The results are show in figure \ref{fig:com1} and figure \ref{fig:com2}. The red line (pq) represents the algorithm in this paper, the blue line represents OmniQuant, the green line represents GPTQ , and the sky blue line represents AWQ, with data sourced from their respective papers. The purple horizontal line represents the perplexity of the original model without any quantization method. Other algorithms run at 3-bit and 4-bit, while PQ runs under the following hyperparameters: (2, 64, 1024), (2, 64, 2048), (2, 64, 4096), (2, 128, 4096), (2, 256, 4096), (3, 64, 1024), (3, 64, 4096), (3, 256, 4096). Some data points with higher quantization bit numbers but higher perplexity were removed to keep the chart concise.

As can be seen, at the precision level where other algorithms require 4 bits, the proposed algorithm can achieve the same precision with fewer bits.

\bibliographystyle{rusnat}
\bibliography{reference}


\end{document}